# Plan3D: Viewpoint and Trajectory Optimization for Aerial Multi-View Stereo Reconstruction

BENJAMIN HEPP, ETH Zurich

MATTHIAS NIESSNER, Technical University of Munich, Stanford University

OTMAR HILLIGES, ETH Zurich

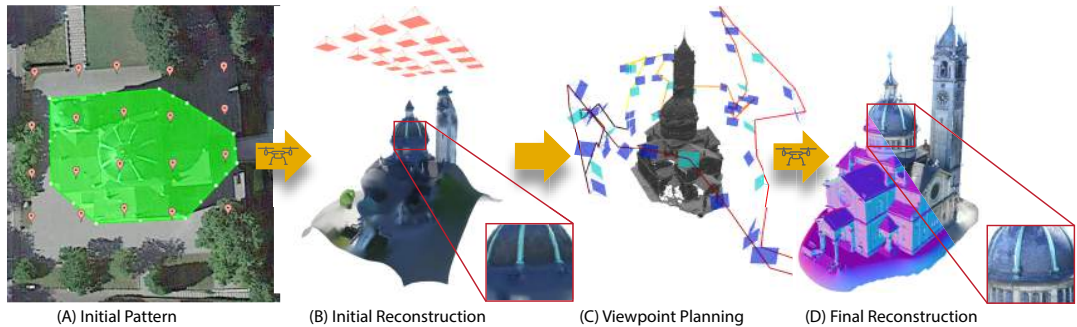

(A) Initial Pattern     (B) Initial Reconstruction     (C) Viewpoint Planning     (D) Final Reconstruction

Fig. 1. We propose an end-to-end system for 3D reconstruction of building-scale scenes with commercially available quadrotors. (A) A user defines the region of interest (green) on a map-based interface and specifies a pattern of viewpoints (orange), flown at a safe altitude. (B) The pattern is traversed and the captured images are processed resulting in an initial reconstruction and occupancy map. (C) We compute a viewpoint path that observes as much of the unknown space as possible adhering to characteristics of a purposeful designed camera model. The viewpoint path is only allowed to pass through known free space and thus the trajectory can be executed fully autonomously. (D) The newly captured images are processed to attain the final high-quality reconstruction of the region of interest. The method is capable of capturing concave areas and fine geometric detail.

We introduce a new method that efficiently computes a set of viewpoints and trajectories for high-quality 3D reconstructions in outdoor environments. Our goal is to automatically explore an unknown area, and obtain a complete 3D scan of a region of interest (e.g., a large building). Images from a commodity RGB camera, mounted on an autonomously navigated quadcopter, are fed into a multi-view stereo reconstruction pipeline that produces high-quality results but is computationally expensive. In this setting, the scanning result is constrained by the restricted flight time of quadcopters. To this end, we introduce a novel optimization strategy that respects these constraints by maximizing the information gain from sparsely-sampled view points while limiting the total travel distance of the quadcopter. At the core of our method lies a hierarchical volumetric representation that allows the algorithm to distinguish between unknown, free, and occupied space. Furthermore, our information gain based formulation leverages this representation to handle occlusions in an efficient manner. In addition to the surface geometry, we utilize the free-space information to avoid obstacles and determine collision-free flight paths. Our tool can be used to specify the region of interest and to plan trajectories. We demonstrate our method by obtaining a number of compelling 3D reconstructions, and provide a thorough quantitative evaluation showing improvement over previous state-of-the-art and regular patterns.

---

## 1 INTRODUCTION

High-quality 3D reconstructions lie at the heart of many applications in computer graphics, AR/VR, robotics and GIS,
architectural and urban planning. Motivated by this need for high-quality 3D models, techniques for the acquisition
of building-scale geometry have rapidly advanced. Even with monocular cameras, 3D reconstructions of impressive
quality can be attained using state-of-the-art multi-view stereo (MVS) methods [Fuhrmann et al. 2014, 2015; Jancosek
and Pajdla 2011; Knapitsch et al. 2017; Langguth et al. 2016; Schönberger and Frahm 2016; Schönberger et al. 2016;
Schöps et al. 2017]. However, the final reconstruction quality depends to a large degree on the availability and *quality*
of the set of input images [Goesele et al. 2007; Seitz et al. 2006] (more is not always better).

   Given the emergence of small and affordable aerial robots (MAVs), equipped with high resolution cameras, it is
a natural choice to leverage these for image acquisition. In fact several commercial applications exist for this task
(e.g., Pix4D [Pix4D 2017] or Agisoft PhotoScan [Agisoft 2014]). Due to safety and absence of detailed environmental
information (e.g., obstacles) such tools revert to flying regular grid-like patterns or circles at a safe overhead distance.
However, the resulting viewpoints are in many cases insufficient for high quality 3D reconstruction: parts of the building
such as walls and convex areas underneath overhangs may be occluded from overhead and images captured from far
away may lack information important for reconstruction of fine details. Furthermore, such acquisition strategies do not
directly account for complete coverage of the scene and do not explicitly reason about the choice of viewpoints with
respect to expected feature matching quality and baseline between viewpoints. Moreover, current MAVs are battery
constrained to 10-15 minute flight time, making intelligent viewpoint selection an even more pressing issue.

   In this paper, we propose an algorithm for the automated planning of viewpoint tours for aerial 3D scanning. We
demonstrate its utility in the context of building-scale dense 3D reconstructions. We pose this problem in a mathematical
optimization framework, based on the objective of *maximizing information* in terms of *observed space* weighted by
uncertainty from as few viewpoints as possible. This objective results in *minimizing unobserved space*. Starting from a
coarse input scan the method selects a small number of viewpoints, each providing additional information about the
environment, and considers constraints in terms of stereo matching, flight time of the MAV and collision freedom. The
resulting plan can be flown by a real quadcopter and the images are processed via a SfM & MVS pipeline [Fuhrmann
et al. 2015; Schönberger and Frahm 2016; Schönberger et al. 2016] to attain high-quality 3D models.

   The above optimization problem involves instances of the coverage set problem for viewpoint selection and the
traveling salesman problem for path planning. Both problems are known to be NP-hard [Chen et al. 2011]. In order to
make the problem computationally tractable, we introduce an approximated camera model to measure the expected
information gain (IG) contribution from an individual viewpoint, independently from all other viewpoints. While this
camera model does not accurately represent monocular viewpoints in MVS settings, it allows us to formulate the
problem with a submodular objective function [Krause and Golovin 2012], making the computation feasible while
giving good approximation bounds. We show that both model and problem approximation yield very good results in practice, and can be effectively combined with an edge selection strategy during trajectory generation that favors camera motion which will lead to good sparse matches and hence good camera pose estimates.

We have developed a simple pipeline that allows novice users to reconstruct building-scale outdoor scenes with little prior knowledge and effort. Based on a user-specified region of interest and no-fly zones a simple overhead pattern is flown for bootstrapping. An initial reconstruction from the collected images is used for planning a sequence of viewpoints through free space and within the maximum travel budget. Finally, the trajectory is flown and a dense 3D model is attained via MVS reconstruction. We have produced a number of reconstructions including a free-standing office building, a historical downtown building in cluttered environment, and an entire renaissance church (see Fig. 1). Furthermore, we evaluate our method quantitatively using a 3D game engine and show that it outperforms both regular patterns and the previous state-of-the art [Roberts et al. 2017].

In summary, we contribute: i) an optimization criteria that is based on volumetrically-represented information content and that maximizes information about uncertain space subject to a travel budget and free-space constraints; ii) a camera-model that makes the optimization problem suitable for a sub-modular formulation, iii) a recursive strategy to approximately solve the sub-modular optimization problem, iv) a quantitative comparison with strong baselines on synthetic scenes and qualitative results on three varied and challenging outdoor scenes.

## 2 RELATED WORK

Our work builds upon a large body of work in the computer graphics and robotics literature. Here we review the most salient work in aerial 3D scanning, path planning and monocular 3D reconstruction.

*3D Reconstruction: SfM and MVS.* The theory behind structure from motion (SfM) and multi-view stereo reconstruction (MVS) methods [Hartley and Zisserman 2003; Triggs et al. 1999] forms the foundation for an impressive diversity of image-based 3D reconstruction approaches. SfM methods can now obtain city-scale reconstructions from unstructured image collections [Agarwal et al. 2010, 2009; Snavely et al. 2006, 2008; Wu et al. 2011] with remarkable levels of quality [Fuhrmann and Goesele 2014; Furukawa and Ponce 2010; Goesele et al. 2006, 2007; Strecha et al. 2008]. Many of these algorithms are now publicly available as open source projects. For example, CMPMVS [Jancosek and Pajdla 2011], MVE [Fuhrmann et al. 2015], SMVS [Langguth et al. 2016], COLMAP [Schönberger and Frahm 2016; Schönberger et al. 2016], or as commercial solutions, such as Pix4D [Pix4D 2017] or Agisoft Photoscan [Agisoft 2014]. A very recent work by Knappitsch et al. [2017] provides an overview of state-of-the-art SfM and MVS methods, and introduces an impressive benchmark dataset along with an evaluation for such approaches. We incorporate domain knowledge about the SfM and MVS process into our planning algorithm but our method is agnostic to the exact underlying SfM and MVS implementation.

*Image Selection for MVS.* It has been shown that the quality and speed of most MVS algorithms significantly depends on the selection of input images and not all input images contribute equally to reconstruction quality, leading to unnecessary processing time if all images are used. Using too many images can even lead to degraded quality in the reconstruction [Seitz et al. 2006]. Most MVS pipelines use relatively simple techniques to select images such as k-nearest images. Several automatic techniques for image selection have been proposed based on heuristics [Farid et al. 1994] or on contours [Kutulakos and Dyer 1992]. More recently, Hornung et al. [2008] proposed a method that incrementally selects images based on maximizing coverage of a proxy geometry. It has also been proposed to leverage viewpoint entropy (similar to IG in active vision) for view selection in image-based rendering [Vázquez et al. 2003].

While not directly focusing on MVS an image selection process based on covariance propagation has been proposed for incremental SfM [Haner and Heyden 2012]. However, all these techniques either expect a complete set of already captured input images or update the 3D reconstruction on-the-fly, making them unsuitable for our settings of finding the best viewpoints for dense aerial 3D reconstruction.

*RBG-D Reconstruction.* In addition to monocular methods, many 3D reconstruction approaches are inspired from range sensing technology. A fundamental component is the volumetric fusion work by Curless and Levoy [Curless and Levoy 1996] whose implicit surface representation is at the core of most prominent real-time RGB-D reconstruction methods, such as Kinect Fusion [Newcombe et al. 2011] and many others [Chen et al. 2013; Dai et al. 2016; Nießner et al. 2013]. Poisson Surface reconstruction is another widely used 3D reconstruction method, where the 3D surface is defined by an energy minimization problem over an implicit, volumetric function [Kazhdan et al. 2006; Kazhdan and Hoppe 2013].

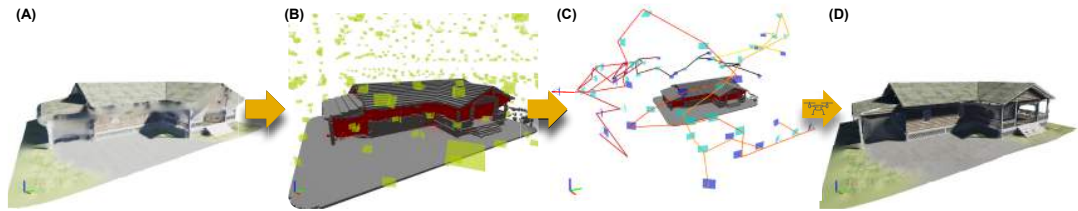

Fig. 2. Overview: (A) The method is initialized via images from a regular overhead pattern (not shown). (B) We generate a set of viewpoint candidates (yellow) and use a submodular optimization formulation to find an optimized viewpoint path maximizing information about uncertain space (voxels in red). (C) The planner generates a free-space trajectory and takes constraints on SfM & MVS reconstruction as preference for fronto-parallel views and matching quality into consideration. Where necessary viewpoints for sparse matching (cyan) are inserted. (D) The final, high quality reconstruction is attained via a SfM & MVS pipeline [Fuhrmann et al. 2015; Schönberger and Frahm 2016; Schönberger et al. 2016].

*Aerial 3D reconstruction.* Several commercial tools aid users in flight planning and processing of aerial imagery for 3D reconstruction [Pix4D 2017; Robotics 2017]. However, these tools only produce regular overhead patterns that have to be flown at a safe obstacle-free altitude. Due to a lack of any prior model information these tools are also incapable of reasoning about coverage and viewpoint quality, limiting the resulting 3D reconstruction completeness and quality. A number of systems leverage MAV-mounted RGB-D sensors to create 3D reconstructions in real-time [Du et al. 2016; Heng et al. 2011; Loianno et al. 2015; Sturm et al. 2013]. However, these focus on the reconstruction method and do not consider the problem of planning trajectories for reconstruction.

*Exploration and active vision.* Planning trajectories through (partially) unknown scenes is a classic problem in the robotics literature (see [Scaramuzza et al. 2014] for a recent review). Frontier-based exploration algorithms [Yamauchi 1997] are often used for autonomous mapping and exploration of the environment using stereo [Fraundorfer et al. 2012], RGB-D, or monocular cameras [Shen et al. 2011]. The goal in our work differs since we are not only interested in a coarse map for navigation but in a dense, high-quality 3D scan of building-scale scenes.

Scanning an a priori unknown scene is a challenging problem due to the lack of available information and absence of useful priors. An interesting approach to guide the recording process was proposed in [Hoppe et al. 2012a] where an online SfM system is combined with measures for redundancy and sampling distance to give live feedback on the expected reconstruction quality to a user. However, in our context we are already given an initial coarse model of the scene and reason about the whole recording sequence in volumetric space before executing it. In active vision approaches, planning of the next view is done on-the-fly by incorporating information from previous scans, an approach known as next-best-view (NBV) and itself an instance of the set coverage problem which is NP-hard. Much work has been dedicated to efficient approximations [Chen et al. 2011] which have been utilized for 3D scanning [Dunn and Frahm 2009; Forster et al. 2014; Kriegel et al. 2015; Wenhardt et al. 2007]. Similar to our formulation such approaches leverage information gain (IG) as optimization quantity. Another interesting approach uses two independent cameras that can collaborate to form a stereo pair [Mendez et al. 2017]. In contrast to our work these approaches either aim for establishing a coarse map for navigation or have not shown high quality dense reconstruction resulting from planned viewpoints.

In graphics, several approaches for quality driven and automated 3D scanning have been proposed including NBV approaches using volumetric [Khalfaoui et al. 2013] and poisson mesh based [Wu et al. 2014] metrics for scanning quality. Fan et al. [2016] propose view- and path-planning algorithms to scan collections of small objects using a structured light scanner. Xu et al. [2015] propose a method for automatic scanning of indoor scenes but focus on guiding the scanning process via object recognition and scene understanding. While these works share our goal of optimizing for reconstruction quality, all of them are designed for structured light scanners, robotic arms, or wheeled robots and comparatively small scenes and hence are not applicable in the context of aerial, monocular (offline) scanning.

In photogrammetry, the task of finding the best viewpoints for accurate 3D scanning is known as the photogrammetric network design problem (NDP). Early work [Fraser 1984] already highlights the difficulties of the problem including high non-linearity and multi-modality which makes it difficult to model in typical optimization frameworks. Model-based approaches [Mason et al. 1997] and such driven by model uncertainty [Olague and Mohr 2002] have been proposed but only deal with the geometric aspect of the task in highly-controlled environments and idealized localization settings (i.e., using fiducial markers) but do not consider the influence of texture and appearance on SfM and MVS pipelines.

The methods in Hoppe et al. [2012b] and Bircher et al. [2015] use a prior mesh to reason about coverage of the scene; however, they only use the mesh for collision checking which might be insufficient for non-watertight meshes with unobserved parts. Their main goal is to find viewpoints that cover the whole scene, whereas we maximize the observed information about a specific object of interest. Once budget constraints like the number of viewpoints or flight-time are introduced such approaches may lead to arbitrarily bad viewpoint selections. The work in [Huang et al. 2017] performs reconstruction online by iteratively capturing a planned sequence of next-best-views and computing depth maps with a fast Multi-View-Stereo method. In contrast our focus is not on planning in an online fashion but on reconstructing a high quality surface model at building scale. The method in [Roberts et al. 2017] is most similar to ours in spirit. The authors also formulate the viewpoint selection as a submodular optimization problem. However, to solve the problem the authors perform two problem approximations resulting in a mixed-integer linear program that is solved with an off-the-shelf solver. Due to these approximations it is not clear whether any approximation bound can be given for the original problem. More importantly, in all our experiments the solver never finds the optimum and it is up to the user to decide how long to run the optimization. For reasonable runtimes of 10 minutes we observe that our approach consistently outperforms this method.

To the best of our knowledge, the view selection and path optimization problem for dense aerial reconstruction has not been formulated in the same way as proposed here and has not been solved in the same efficient manner as we do.

## 3  SYSTEM OVERVIEW

The aim of our method is to automate high-quality 3D scanning of building-scale scenes using commodity quadrotors equipped with a single RGB camera. To this end, we have developed a simple pipeline, illustrated in Fig. 1 and 2, which is built on a mathematical optimization formulation for viewpoint and trajectory planning. In the following, we provide an overview of the workflow from a user's perspective and then detail the planning algorithm.

First, a user defines a region of interest (ROI) and specifies a simple and safe overhead pattern via a map-based interface to acquire an initial set of images. In urban environments, it is also possible to specify no-fly zones in order to avoid other buildings, powerlines, and to adhere to airspace regulations. The quadrotor flies this regular pattern and records an initial set of images. These recordings are then processed via a state-of-the-art SfM and MVS pipeline by Schönberger et al. [2016; 2016] to attain camera poses together with depth and normal maps for each viewpoint. To generate a 3D surface reconstruction, the depth maps are fused into a dense point cloud, and utilizing the Poisson Surface Reconstruction method [Kazhdan and Hoppe 2013] a mesh is extracted (Fig. 2, A).

It is important to note that this initial reconstruction is highly inaccurate and incomplete since the viewpoints stem from a simple, regular pattern, flown at relatively high altitude to avoid collisions.

In addition to the initial reconstruction, we compute a volumetric occupancy map containing occupied, free-space, and unobserved voxels. Each voxel also carries with it a measure of observation quality. We describe this occupancy map in more detail at the end of this section. Our occupancy map is based on the implementation from [Hornung et al. 2013].

The occupancy map (Fig. 2, B) is used during planning to reason about free-space and collision freedom as well as approximation of the observable surface area from any given viewpoint and the (remaining) uncertainty about the scene. The main objective of our optimization formulation is to maximize total information (i.e. certainty about voxels in the region of interest) while staying within the travel budget of the quadrotor and respecting constraints imposed by SfM and MVS. Intuitively, the total information corresponds to the observation count of all voxels within the region of interest. In theory, given lambertian surfaces, once all voxels have been observed multiple times (from different non-grazing angles) the entire surface can be reconstructed with high quality. This goal has to be traded-off with limited battery time of the robot and computational cost as the evaluation of all possible viewpoints during planning is clearly infeasible.

Furthermore, it has been shown that at some point adding views yields diminishing returns [Hornung et al. 2008; Seitz et al. 2006; Waechter et al. 2014]. Therefore, we propose an efficient way to generate and evaluate viewpoint candidates (see Fig. 2, B) during planning, alongside a method to find a sequence of 'good' viewpoints to be flown in order to maximize reconstruction quality.

Fig. 2, C shows the output of our planning method, where viewpoints that were added due to their contributed information are rendered in blue. Additional viewpoints that were added to ensure that the SfM & MVS backend can register all images into a single reconstruction are rendered in cyan. The edges are color-coded to signal MAV progress along the path. The plan is then executed by the drone and the acquired images are used to update the 3D model (Fig. 2, D). Note how convex areas underneath the front-porch and garage roofs have been carved and how overall detail has been increased. However, there are still parts of the scene that have not been reconstructed properly. This is due to approximations simplifying assumptions in our formulation. We discuss these limitations in 6.

*Volumetric Occupancy Map.* The occupancy map $OM$ is essential in distinguishing between occupied, free and unobserved space. This is encoded by an occupancy value $oc(\tau) \in [0, 1]$ for each voxel $\tau \in OM$. Here we overload the term occupancy to encompass both a known occupancy and an unknown occupancy, i.e. a value close to 0 encodes a known empty voxel, a value close to 1 encodes a known occupied voxel and a value close to 0.5 encodes an unknown voxel.

We initialize the occupancy map with occupancies of $\sigma = 0.5$ (i.e. unknown). From each viewpoint, we cast a ray through the center of each pixel until we reach the 3D point given by the depth value [Curless and Levoy 1996; Hornung et al. 2013]. The occupancy of each traversed voxel is updated according to an inverse sensor model (see Sec.7 for more details). The size of a voxel is $vs = 0.2m$ in all our experiments. We denote the position of a voxel $\tau$ as $\tau.\mathbf{p}$. In the following we only consider voxels inside the region of interest when computing the viewpoint information.

## 4 METHOD

In this section, we discuss our optimization framework to obtain viewpoint trajectories for 3D scanning. One difficulty that presents itself is that the final objective of 3D reconstruction quality cannot be measured directly due to the absence of ground-truth data and the offline nature of the SfM and MVS pipeline. Hence, we require a surrogate quantity that can be used for optimization. At the core of our method lies an objective function based on the total information gained by the collection of viewpoints along a computed trajectory. We first introduce the formal optimization problem and then discuss how an approximate camera model can be leveraged to modify the original problem to allow for efficient maximization.

### 4.1 Optimizing viewpoint trajectories

The high-level goal is to find an optimized subset of viewpoints, from a larger set of candidate views, that maximizes the information gained about the 3D surface of the scene.

We assume that we are given a graph $G = (C, M)$ of viewpoint candidates $C$ alongside observed voxels and motions $M$ between viewpoints as edges. Each viewpoint $v \in C$ has an associated position and orientation denoted as $v.\mathbf{p}$ and $v.\mathbf{q}$. We require each of the motions in $M$ to be collision-free and that the connected viewpoints can be matched later on in the MVS pipeline. We detail the construction of this graph in Sec. 4.4.

The goal of the method is to generate a trajectory (i.e. a path through a subset of the nodes in the candidate graph) for the quadcopter that yields good reconstruction quality and fulfills robot constraints . In our case we are given a maximum travel distance of $L_{max}$. Let $VP = (vp_1, \ldots, vp_n)$ be the sequence of viewpoints to be traversed during image capture where $vp_i \in C$. We denote with $L(VP)$ the geometric length of the trajectory $VP$ and with $S$ the set of all sequences $VP$.

Formally, we want to solve the following optimization problem:

$$VP^* = \underset{VP \in S}{\operatorname{argmax}} \; I(VP) \tag{1}$$

$$\text{such that } L(VP) \leq L_{max} \quad,$$

where $I(VP)$ is our objective function that measures the amount of information contributed by the respective viewpoints. We can write the objective function as a sum over the information of each non-free-space voxel

$$I(VP) = \sum_{\tau \in OM \setminus OM_{free}} VI(\tau, VP) \tag{2}$$

$$OM_{free} = \{\tau \in OM : oc(\tau) \leq oc_{free}\} \quad, \tag{3}$$

where $VI(\tau, VP)$ is our camera measurement model and specifies how much information of voxel $\tau$ is contributed by the traversed viewpoints $VP$ (details in Sec. 4.2) and $oc_{free}$ is a lower threshold that determines when a voxel is considered to be free space (see Appendix for details).

Solving Eq. (1) is in general prohibitively expensive for non-trivial real world problems [Krause and Golovin 2012] since the total information of a voxel depends on the set of all traversed viewpoints. As our viewpoint graphs contain $5,000 - 10,000$ viewpoints it is computationally infeasible to enumerate all these viewpoint sets to find the best one.

### 4.2 Submodular voxel information

To make the problem tractable, we approximate the contributed information of a voxel by assuming that a single viewpoint $v$ can directly provide information about the 3D surface, i.e., for a single voxel $\tau$ and viewpoint $v$ the contributed information can be written as $vi(\tau, v)$. We do this by incorporating terms that encourage close-up and fronto-parallel views. The computation of the contributed information is detailed in Sec. 4.4 (see Eq. (7)) and depends on the incidence angle of the observation ray and the normal of the voxel.

Clearly, this model is far from reality since it entirely ignores the stereo-matching process and does not explicitly encourage the selection of images from diverse viewpoints. However, it does allow us to re-formulate our optimization problem in a way that it exploits sub-modularity in the objective function and therefore allows for more efficient maximization of the problem [Krause and Golovin 2012]. Note that to explicitly incorporate stereo matching into the objective function we would want a view to give a high incremental objective value if another view can be matched to it. I.e. let $A, B \in 2^S$, $A \subset B$ and $x \in S \setminus B$ where $S$ is the set of all possible camera poses. Lets assume that $B$ contains a view that allows good stereo matching with $x$ while $A$ contains no such view. To incorporate stereo matching into the objective function $I : 2^S \rightarrow \mathcal{R}$ we would like to have $I(B \cup x) - I(B) \gg I(A \cup x) - I(A)$. We see that such a function is not submodular as it would violate the submodularity condition $I(B \cup x) - I(B) \leq I(A \cup x) - I(A)$.

We can finally write the total information of a voxel $\tau$ contributed by a set of viewpoints as

$$VI(\tau, VP) = \min\left(1, \sum_{v \in VP} vi(\tau, v)\right) \quad, \tag{4}$$

where the *min* ensures that the total information for each voxel saturates at 1. It is easy to see that this objective function is submodular [Krause and Golovin 2012] by writing the information gain resulting from adding $v$ to the viewpoint sequence $VP$:

$$IG(\tau, v, VP) = VI(\tau, \{v\} \cup VP) - VI(\tau, VP) \tag{5}$$

$$= \min\left(vi(\tau, v), 1 - VI(\tau, VP)\right) \quad. \tag{6}$$

We note that $VI(\tau, VP_A) \leq VI(\tau, VP_B)$ for $VP_A \subseteq VP_B$ and thus the submodular property $IG(\tau, v, VP_A) \geq IG(\tau, v, VP_B)$ is fulfilled.

As mentioned above, our submodular objective function does not directly incorporate stereo matching so we would like to encourage the optimization to select multiple views of the same voxel to account for the requirements of stereo matching. To this end we reduce the contributed information per viewpoint by a discount factor $\xi$. This encourages on average $1/\xi$ cameras to observe the same point, leading to good MVS reconstructions in practice. Please also note that our formulation continues to enforce stereo matching constraints via the edge selection strategy discussed in Sec. 4.4.

### 4.3 Maximizing the submodular formulation

Our camera model exposes desirable structure in the optimization problem Eq. (1). Since the individual viewpoints now provide diminishing returns given other additional viewpoints and more viewpoints can never reduce the total information, the objective function $I(VP)$ is both *monotone* and *submodular*. While the problem is still NP-complete in general, submodularity provides guarantees on the approximation quality of a greedy algorithm for the when we are allowed to select a fixed number of viewpoints [Krause and Golovin 2012]. This guarantee does not hold anymore when we introduce a travel/time budget constraint on the path through the selected viewpoints. By combining the greedy algorithm and the cost-benefit algorithm [Wolsey 1982; Zhang and Vorobeychik 2016] and choosing the better solution we are guaranteed to be within $(1 - 1/e)/2 \simeq 0.32$ of the optimal solution. However, the cost-benefit algorithm requires $|C|$ evaluations of the contributed information $vi(\tau, VP)$ for each added viewpoint and thus does not scale well with an increasing set of viewpoint candidates and an increasing travel/time budget. The algorithm proposed in [Chekuri and Pal 2005] also solves the submodular optimization problem with a travel-budget constraint. The algorithm fixes a start and end viewpoint and follows a recursive strategy by selecting a *middle* viewpoint and splitting the trajectory into a first (start to middle) and second part (middle to end). The travel-budget is also split into a portion for the first and second part. The first and second part are then computed by taking a recursive step. Interestingly, the authors are able to provide an approximation ratio of $\log O*$ for this algorithm where $O*$ is the optimal value of the objective function. However, the algorithm is severely restricted in practice as it enumerates all possible *middle* nodes and all possible splits of the travel-budget (which is assumed to be integer valued) for each recursion step. Also note that the objective and the contributed information of each viewpoint are assumed to be integer valued (otherwise we could just scale all information values to achieve a better approximation ratio). In our setting with real-valued budgets and travel costs and a high number of viewpoint candidates this algorithm is not feasible.

We introduce a practical adaptation of the method in [Chekuri and Pal 2005] to solve Eq. (1). In all our experiments this method performs favorably and always achieves a better solution than the greedy method. Note however, that our solution is only an approximation as the problem is still NP-complete in general.

Our approach can be seen as striking a middle-ground between the greedy algorithm and the cost-benefit algorithm. When observing the greedy algorithm one usually sees that the next viewpoint being picked is very far away from already selected viewpoints. This makes sense as viewpoints far away will have little overlap with already selected viewpoints and there is no penalty for large distances between selected viewpoints. This often leads to suboptimal selections as much of the viewpoint budget is used up very early and later on viewpoints can only be selected very close to the travel paths between the earlier viewpoints. On the other hand, the cost-benefit algorithm usually ends up selecting viewpoints that are very close to existing viewpoints due to the penalty of choosing viewpoints with large distances to the already selected ones. This often results in clusters of viewpoints with low coverage. Our method will typically also pick new viewpoints that are far away from existing viewpoints as we do not directly penalize distances in the selection. However, after selecting a viewpoint we split the budget into a first part (before reaching that viewpoint) and a second part (after reaching that viewpoint). We then recursively continue selecting further viewpoints in the first part and only later continue with the second part.

Our method proceeds as follows: the budget is split into a first and second part and a middle node (viewpoint) is selected. A recursion step is made for the first part and the second part. In contrast to [Chekuri and Pal 2005] we only perform an equal split of the budget for the first and second part and select as the middle node the viewpoint with the highest information gain that is reachable with the current budget. The recursion naturally ends when no new

 

viewpoint can be reached with the available budget. Additionally, we adjust the budget for the recursion of the second part to make use of all remaining budget after the recursion of the first part has finished. A formal description of the method is given in Alg. 1.

---

**Procedure** *recursiveGreedy* **is**
    **Input:** $V, V_s, V_e, B$
    **Output:** $VP$
    $V_m \leftarrow$ Viewpoint with maximum information gain that is still reachable with budget $B$.
    **if** $V_m == \emptyset$ **then**
        | Return ()
    **end**
    $VP_1 \leftarrow$ *recursiveGreedy*( $V \cup \{V_m\}, V_s, V_m, B/2$)
    $B_1 \leftarrow$ Compute travel length of $VP_1$
    $B_2 \leftarrow B - B_1$
    $VP_2 \leftarrow$ *recursiveGreedy*( $V \cup VP_1, V_m, V_e, B_2$)
    $VP \leftarrow (V_s) + VP_1 + (V_m) + VP_2 + (V_e)$
**end**

**Algorithm 1:** Recursive greedy algorithm to maximize the optimization problem in Eq. (1). The recursive procedure takes the set of viewpoints $V$ currently on the path $VP$ (i.e. $V = \{v \; \forall \; v \in VP\}$) and the start viewpoint $V_s$, end viewpoint $V_e$ and budget $B$ of subproblem as input. It returns the viewpoint path for the subproblem. The initial call of the procedure is *recursiveGreedy*$(\emptyset, V_i, V_i, B_{total})$, where $V_i$ is the viewpoint with the overall maximum score and $B_{total}$ is the full travel budget of the quadrotor. The middle viewpoint $V_m$ can be computed in an efficient manner by keeping a sorted list of IG and evaluating them in a lazy fashion (see Appendix for more details). A viewpoint is reachable with the current travel budget if the travel distance from $V_s$ to $V_m$ and from $V_m$ to $V_e$ does not exceed the budget $B$.

### 4.4 Viewpoint candidate graph

In this section we detail the generation of the candidate viewpoint graph including computation of contributed information per viewpoint and generation of free-space motion paths. The candidate viewpoint graph $G$ consists of a large number of candidate viewpoints and the corresponding observed voxels as nodes and free space motions between viewpoints as edges. Note that we first generate the set of candidate viewpoints and afterwards compute motion paths between them.

Ideally, the set of candidate viewpoints should consist of all camera poses that are useful for reconstructing the surface in the region of interest. At the same time the set should be as small as possible to enable faster computation on the set. We note that a pair of images with a fixed baseline will be visually more similar if the images are taken far away from the region of interest. Thus the candidate generation process (see Alg. 2 in Appendix) is designed to perform a denser sampling of viewpoints inside or in the vicinity of the region of interest and a less dense sampling for viewpoints far away from the region of interest. The set of candidate viewpoints is seeded either by the manually specified viewpoints or those stemming from an earlier iteration of the planning algorithm. These initial candidates are also added to an exploration queue. To sample new 3D candidate positions we take the first 3D position from the exploration queue and generate 6 new positions by adding an offset in the $-x, +x, -y, +y, -z$ and $+z$ direction respectively. The resulting positions are discarded if they are too close to existing viewpoint candidates or do not lie in free space, otherwise they are added to $C$ and to the exploration queue. The orientation of new viewpoints is determined by random sampling with a bias towards the region of interest.

A viewpoint is considered to lie in free space if the occupancy of all voxels intersecting with the drone's bounding box is below the threshold $oc_{free}$ (see Appendix for more details).

We furthermore require the camera orientation to fall within the limits of the physical camera (i.e., camera roll $\phi = 0$ and pitch $\theta \in [-\pi/2, 0]$). We run this procedure until no more new viewpoints can be added or we reach a maximum viewpoint budget. For typical scenes, $5,000$ - $10,000$ viewpoint candidates are generated (see Fig. 2, B).

Importantly $G$ is not arranged in a uniform grid. Instead we let the sampling offset between neighboring viewpoints grow with increasing distance from the region of interest. This prevents an explosion in the number of viewpoints for larger scenes.

**Viewpoint information**

After generating viewpoint candidates, we compute the visible voxels for each viewpoint needed to evaluate the corresponding contributed information by ray-casting into the occupancy map. Compared to a simple rendering of the coarse Poisson mesh from the initial scan the ray-cast ensures that we handle occlusions by objects that were not reconstructed (i.e. unknown voxels). Note that this operation is fairly expensive but only has to be performed once for each viewpoint candidate.

The information $vi(\tau, v)$ contributed by observing a voxel $\tau$ with viewpoint $vp$ is then given by

$$vi(\tau, v) = \frac{1}{\xi} vi_i(\tau, v) \, vi_r(\tau, v) \qquad (7)$$

$$vi_i(\tau, v) = \exp(-\beta_i^F \max(\gamma - \beta_i^T, 0^\circ))$$

$$vi_r(\tau, v) = \exp(-\beta_r^F \max(px(\tau, v) - \beta_r^T, 0px)) \quad,$$

where $vi_i(\tau, v)$ and $vi_r(\tau, v)$ are incidence- and resolution-dependent factors respectively, $\gamma = \arccos((v.\mathbf{p} - \tau.\mathbf{p}) \cdot n(\tau, v))$ is the angle between the incident viewing ray and the surface normal (extracted from the mesh) and $px(\tau, v) = \frac{f \, vs}{d(\tau.\mathbf{p}, v.\mathbf{p})}$ is the number of pixels that a fronto-parallel voxel stretches if projected onto the viewpoint's image plane with a focal length $f$. Note that $d : \mathbb{R}^3 \times \mathbb{R}^3 \rightarrow \mathbb{R}$ refers to the euclidean distance. Both factors have the same functional form: a constant value of 1 up to a certain threshold $\beta_i^T, \beta_r^T$ followed by an exponential decrease with falloff factors $\beta_i^F, \beta_r^F$. We empirically determined the following values to work well on a number of scenes: $\beta_i^T = 25^\circ$, $\beta_i^F = \frac{1}{25^\circ}$, $\beta_r^T = 6px$, $\beta_r^F = \frac{1}{3px}$. These values are used for all our experiments.

Note that we compute the normal of a voxel in a per-viewpoint fashion. This is accomplished by rendering the Poisson mesh of the initial reconstruction from the corresponding viewpoint for which we are computing the voxel information. Furthermore, we consider the voxel's projected size to account for the physical camera resolution. Clearly, the voxel information $vi$ depends on the quality of the initial Poisson mesh. However, as our incidence factor $vi_i$ shows a flat response around an incidence angle of $0^\circ$ (i.e. looking parallel to the surface normal) the voxel information $vi$ is only slightly effected by small amount of noise in the normals of the Poisson mesh. To handle normals that are close to creases of the surface, we compare the distance of camera to mesh and camera to voxel. If these distances differ too much ($\geq 0.5m$) we assume an unknown normal and set the incidence factor $vi_i(\tau, v)$ to 0.2. This incentivizes multiple observations of the voxel for these ambiguous situations. An alternative to rendering the Poisson mesh from each viewpoint would be to compute a single view-independent normal for each voxel by searching for the nearest triangle in the Poisson mesh. However, this has certain limitations such as forcing a single normal for a voxel that is part of a crease of the surface. With our view-dependent normal computation the voxel can have different normals depending from which side it is viewed.

To accelerate the computation of the contributed information by a viewpoint we perform the ray-casting operation on a GPU using an image plane with a resolution of $600 \times 450$ and using a focal length of $f = 345\text{px}$, equivalent to that of the physical camera. We assume rays have an infinitesimal width and shoot a ray through the center of each pixel of the image plane. In our current setting a single voxel with edge-length of $0.2m$ projects onto one pixel at a distance of $\frac{0.2m \cdot f}{1\text{px}} \simeq 70m$.

**Free-space motion paths**

The final step in computing the viewpoint candidate graph $G$ is to connect nodes via traversable edges (see Alg. 3 in Appendix). Traversable refers to a path along which the drone's bounding box does not intersect with any non-free voxel. To this end we move the bounding box along the path with a maximum step-length of $l$, bounding the violation of the obstacle check to a maximum distance of $\sqrt{3}l$. By choosing the bounding box dimensions appropriately, we can ensure, up to the limits of the initial reconstruction, that the path is obstacle free and can be flown safely. For computational efficiency, we first attempt to connect viewpoints via straight lines. If an obstacle is encountered, we employ the rapidly exploring random trees (RRT*) algorithm [Karaman and Frazzoli 2011] to find a free-space motion consisting of piecewise linear segments. A voxel is determined to be in free space if its occupancy is below the threshold $oc_{free}$ (see Appendix for more details).

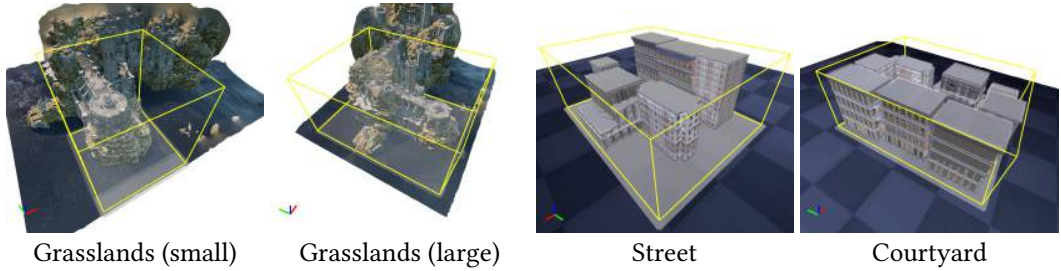

|  Grasslands (small)  |  Grasslands (large)  |  Street  |  Courtyard  |

Fig. 3. Overview of synthetic scenes used for evaluation. Shown in yellow is the bounding box of the region of interest together with the transparent voxel representation and the mesh representation of the scene. The *Grasslands* scenes on the left demonstrates interesting geometry that is mostly observable from easily accessible viewpoints. In contrast the *Street* and *Courtyard* scenes contain simpler geometry but many surfaces can only be observed from viewpoints between the buildings which are difficult to access.

**Sparse Matching**

On most scenes, and inline with findings from the MVS literature, we attain high quality dense depth maps from relatively few images. However, in building-scale scenes clusters of views can be very far apart, making global registration of depth maps impossible. In order to account for this issue, we furthermore include a heuristic that will introduce additional viewpoints for registration and naturally integrates into our viewpoint candidate graph. To this end, we define a criterion based on the visible voxels in each viewpoint. To speed up the computation, we render the occupancy map at a lower resolution and count the voxels $obs(v)$ visible from a viewpoint $v$. Two viewpoints $v_1$ and $v_2$ are then called *matchable* if:

$$|obs(v_1) \cap obs(v_2)| \geq \alpha(|obs(v_1)| + |obs(v_2)|)/2 \quad , \tag{8}$$

ensuring that there is enough overlap of visible voxels between two viewpoints. We only insert motions into the viewpoint graph $G$ if the two connected viewpoints are *matchable*. Incorporating this matching heuristic has proven to be very effective in our synthetic and real-world experiments. In all our experiments, we use a value of $\alpha = 0.4$. Once a matchable motion path is found we add an edge with a weight equal to the motion distance to the graph $G$.

## 5 RESULTS

In this section, we discuss experiments conducted to evaluate our method both quantitatively and qualitatively, on real and synthetic data. Evaluating methods for building-scale robotic 3D reconstruction poses many significant challenges. First, comparative evaluation on *real* scenes is challenging due to changing conditions such as lighting, weather conditions, surrounding objects and modifications of the object itself that can occur over the course of hours. Furthermore, groundtruth information is typically not available for building size scenes. Therefore we report qualitative results from three challenging real-world examples, a renaissance *church* scene, an *office* building and a *historic* building. We also report quantitative results from a thorough evaluation on *synthetic* scenes from a state of the art rendering engine[1].

The method proposed in here consists of many parts such as the robotic platform itself, the planning algorithm and the SfM/MVS pipeline. To unpack the influence of our core contribution (the planning algorithm) we compare our method with several strong baselines, including a state-of-the art method [Roberts et al. 2017], in an ablative manner. First, we repeat the experiment in [Roberts et al. 2017] to compare results from the end-to-end reconstruction of the full system *including* the SfM/MVS pipeline. Next we evaluate the relative performance to [Roberts et al. 2017] with respect to the main optimization objective of achieved total viewpoint information (score). Finally, we compare our method with several strong baselines using reconstructions from recorded depth images to remove the influence of the SfM/MVS pipeline.

### 5.1 Synthetic scenes

We evaluate our method on four *synthetic* scenes stemming from the *Infinity Blade: Grasslands* environment[2] and the *Urban City* environment[3] from the Unreal Engine Marketplace. In the Grasslands environment we use a small region of interest as used in [Roberts et al. 2017] and a large region of interest. The scenes from the Urban City environment are blocks of houses where the inner facades of the building are difficult to observe. An overview of the scenes and the regions of interest is shown in Fig. 3.

To evaluate our algorithm's performance when reconstructing a dense point cloud we closely follow the procedure established in [Knapitsch et al. 2017]. We resample the ground truth mesh until the area of each triangle is below a certain threshold ($0.5m \times 0.5m$). From each triangle 100 points are sampled and the resulting point cloud is resampled on a voxel grid with a voxel size of $vs/2 = 0.05m$ by taking the mean of the points found within the same voxel. The resulting point cloud represents our ground truth. The reconstructed point cloud is resampled on the same voxel grid. To compare both point clouds we compute two quantities, the precision $P$ and the recall $R$, where precision quantifies how many reconstructed points are close to a ground truth point and recall quantifies how many ground truth points are close to a reconstructed point. A point is close to another point if their distance is less or equal to $\delta = 0.1m$. Both

---

 

quantities can be combined in the F1 score $F = 2\frac{P \cdot R}{P + R}$ which is a common performance measure for binary classification. We refer the reader to [Knapitsch et al. 2017] for more details.

### 5.2 Comparison with Roberts et al.

Here we directly compare our reconstruction results with our reimplementation of [Roberts et al. 2017] Like the authors we use Gurobi[4] to solve the resulting mixed-integer linear program and allow a runtime of 10 minutes. We observe that running Gurobi for much longer times does not improve the resulting score significantly (see Appendix 9.3). We show reconstruction results using ground-truth depth maps *and* end-to-end results when performing dense reconstruction using Multi-View-Environment (MVE) [Fuhrmann et al. 2015].

We perform the comparison on the same *Grasslands (small)* scene as in [Roberts et al. 2017] and allow a travel budget of 900m. [Fuhrmann et al. 2015]  We use 20 images arranged in a circular pattern to compute the initial reconstruction. [Roberts et al. 2017]  As shown in Table 1 our method improves upon [Roberts et al. 2017] in both experimental settings. In Fig. 4 we qualitatively compare the reconstruction results using the end-to-end dense reconstruction, showing that our method can recover more surface details.

Grasslands (small)

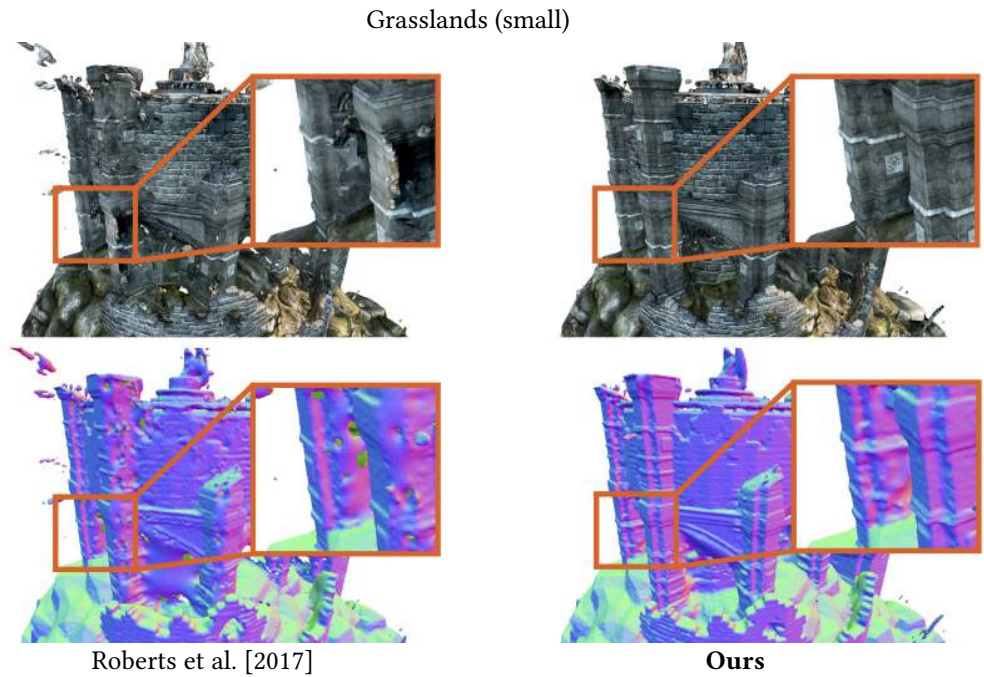

Roberts et al. [2017]                                          **Ours**

Fig. 4. Qualitative comparison of [Roberts et al. 2017] and our method on an end-to-end dense reconstruction of the Grasslands (small) scene. Here we included the images from the initial scan when performing reconstruction. The first row shows the geometry without color whereas the second row shows the texture mapped reconstruction. The method of [Roberts et al. 2017] generates a viewpoint path that fails to capture some geometry information. Note that the missing geometry information also leads to distorted or erroneous textures. (Best viewed in color.)

---

[4]http://www.gurobi.com/

Using ground-truth depth maps

| Method | Precision | Recall | F-Score |
|---|---|---|---|
| Roberts et al. [2017] | 97.22 | 62.53 | 76.11 |
| **Ours** | 96.56 | 67.16 | **79.22** |

Using full SfM/MVS pipeline

| Method | Precision | Recall | F-Score |
|---|---|---|---|
| Roberts et al. [2017] | 81.83 | 64.91 | 72.39 |
| **Ours** | 80.29 | 72.17 | **76.02** |

Table 1. Quantitative comparison of the final reconstruction on the synthetic scene *Grasslands (small)* with a small region of interest and a travel budget of 900*m*. Here we included the images from the initial scan when performing reconstruction. The best F-Score value is highlighted in bold.

## 5.3 Viewpoint score comparison

Here we directly compare our optimization method with the approach in [Roberts et al. 2017] using the same underlying objective function. Recall that the optimization objective is a submodular viewpoint score describing how much surface of a scene has been covered and the solution space is restricted by a budget constraint. We use the *Grasslands (small)* scene and the *Grasslands (large)* scene and measure the achieved viewpoint score for different travel budgets. We ran the method from [Roberts et al. 2017] for 10*min* while our method took less than 10*min* in all cases.

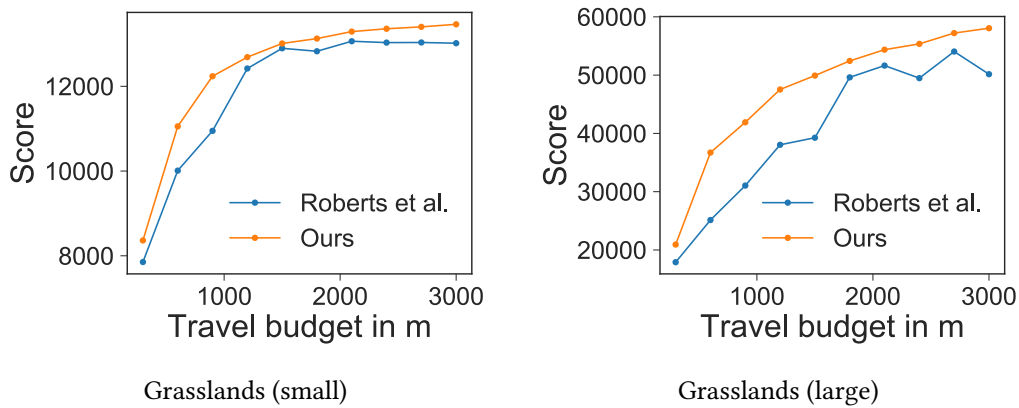

Grasslands (small)                                    Grasslands (large)

Fig. 5. Comparison of submodular optimization methods on the synthetic scenes *Grasslands (small)* and *Grasslands (large)* using the same objective function. Note that we would except both methods to saturate to the same value for an ever increasing budget. Thus we are in particular interested in the performance at intermediate budgets. As can be seen our method performs better for both scenes and all budgets but the improvement is much more pronounced in the larger scenes with more viewpoint candidates.
.

The results in Fig. 5 show that our method consistently achieves a higher viewpoint score compared to [Roberts et al. 2017]. In particular for larger scenes  we can observe an increased difference in scores, suggesting that our method scales better to larger scenes. This is supported by later experiments on different scenes where Gurobi was unable to find a solution within a runtime of 60 minutes unless we reduced the number of considered viewpoints.

### 5.4 Comparison with regular baseline patterns

We compare reconstruction results against strong baselines, including regular patterns such as circles, meanders and hemispheres of different size and our reimplementation of [Roberts et al. 2017]. Note that the hemisphere pattern is computed based on our knowledge of free-space and our viewpoint graph and as such is more advanced than patterns used in current commercial tools. For the simpler baseline patterns (i.e. circle and meander) we choose the number of images to record for the 3D reconstruction in a best practice manner. In the Appendix 9.3 we show that including additional images does not necessarily improve and indeed often degrades the reconstruction performance.

We use three different scenes, *Grasslands (large)*, *Street* and *Courtyard*. For the *Grasslands (large)* scene we allow a travel budget of $1500m$ and use a circle pattern with 20 viewpoints for the initial reconstruction. For the *Street* and *Courtyard* scenes we allow a travel budget of $2700m$ and use 30 images arranged in a meander pattern so that our volumetric mapping can carve away more space between the buildings. Unfortunately, these scenes have repetitive textures which lead to many erroneous artifacts when we run the MVS pipeline end-to-end which make a comparison meaningless. To mitigate this issue we use rendered depth images along the computed viewpoint path and fuse them into a dense point cloud by restricting the maximum distance, incidence angle and requiring at least 3 nearby depth measurements to reconstruct a point. We compute the precision, recall and F-Score on the resulting dense point cloud as described above.

In Fig. 6 we show qualitative comparison of the reconstructions. The measured quantities are shown in Table 2, Table 3 and Table 4. Our method yields higher F-scores than all other methods. The advantage of our method compared to a relatively simple hemisphere pattern becomes apparent in the more realistic *Street* and *Courtyard* scenes where a hemisphere pattern can not provide viewpoints that cover the inner facades of the buildings as they are shielded by the opposite buildings. Our method in contrast puts viewpoints within the free space between the buildings and also surrounding them to cover all surfaces. This is reflected in the higher recall and F-score compared to the hemisphere patterns.

| Method | Precision | Recall | F-Score |
|---|---|---|---|
| Small circle ($35m$ radius) | 77.29 | 13.26 | 22.64 |
| Large circle ($70m$ radius) | 72.55 | 3.92 | 7.44 |
| Small meander ($70m \times 70m$) | 44.68 | 20.00 | 27.64 |
| Large meander ($140m \times 140m$) | 43.60 | 20.57 | 27.95 |
| Small hemisphere ($60m$ radius) | 85.64 | 51.40 | 64.24 |
| Large hemisphere ($75m$ radius) | 81.49 | 50.41 | 62.29 |
| NextBestView / Greedy | 90.63 | 44.56 | 59.74 |
| Roberts et al. [2017] | 87.87 | 50.03 | 63.76 |
| **Ours** | 90.57 | 57.70 | **70.49** |

Table 2. Quantitative comparison of the final reconstruction on the synthetic scene *Grasslands (large)* with a large region of interest and a travel budget of $1500m$. Here we included the images from the initial scan when performing reconstruction. The best F-Score value is highlighted in bold.

### 5.5 Real scenes

In this section we discuss results acquired from real scenes and give a qualitative comparison with baselines from regular patterns as used in commercial tools. All reconstructed models shown here were computed with COLMAP

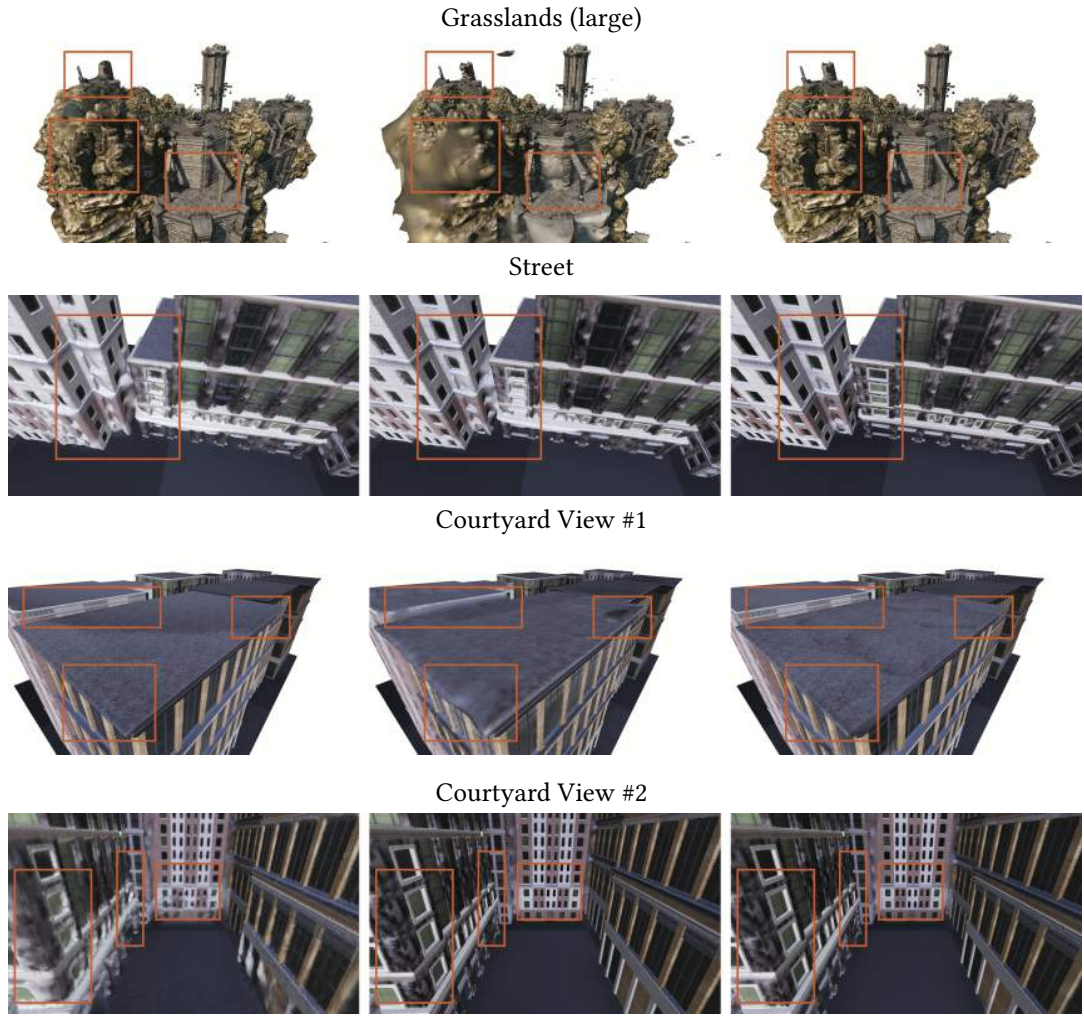

Fig. 6. Qualitative comparison of reconstructed meshes. For the Grasslands (large) scene we show results from a hemisphere pattern, [Roberts et al. 2017] and our method. Here we included the images from the initial scan when performing reconstruction. Both the hemisphere pattern and [Roberts et al. 2017] have missing points or lower point densities in some regions leading to bulging in the computed Poisson mesh. In contrast the reconstruction for our method shows much less bulging. For the *Courtyard* and *Street* scene we show results from a hemisphere pattern, the approach in [Roberts et al. 2017] and our method. Note that our method has more detailed texture and geometry in the lower facades of the building which requires viewpoints that move close enough but not exceeding the overall travel budget. (Best viewed in color and electronically.)

[Schönberger and Frahm 2016; Schönberger et al. 2016]. which showed subjectively better results than MVE [Fuhrmann et al. 2015].

| Method | Precision | Recall | F-Score |
|---|---|---|---|
| Small circle (50$m$ radius) | 86.98 | 15.88 | 26.86 |
| Large circle (75$m$ radius) | 83.21 | 6.50 | 12.06 |
| Small meander (85$m$ × 100$m$) | 61.48 | 11.34 | 19.15 |
| Large meander (100$m$ × 115$m$) | 66.09 | 10.60 | 18.26 |
| Small hemisphere (60$m$ radius) | 91.34 | 63.95 | 75.23 |
| Large hemisphere (75$m$ radius) | 90.06 | 42.35 | 57.61 |
| NextBestView / Greedy | 94.23 | 68.46 | 79.30 |
| Roberts et al. [2017] | 93.60 | 70.42 | 80.37 |
| **Ours** | 93.92 | 71.69 | **81.31** |

Table 3. Quantitative comparison of the final reconstruction on the synthetic scene *Courtyard* with a travel budget of 2700 m. Here we included the images from the initial scan when performing reconstruction. The best F-Score value is highlighted in bold.
Note that we had to limit the number of viewpoints used in our implementation of Roberts et al. [2017] to 3000. Otherwise Gurobi was not able to find any solution within a runtime of 60 minutes.

| Method | Precision | Recall | F-Score |
|---|---|---|---|
| Small circle (50$m$ radius) | 84.37 | 16.91 | 28.17 |
| Large circle (75$m$ radius) | 77.96 | 6.06 | 11.24 |
| Small meander (85$m$ × 100$m$) | 56.94 | 12.01 | 19.83 |
| Large meander (100$m$ × 115$m$) | 60.38 | 11.15 | 18.82 |
| Small hemisphere (60$m$ radius) | 90.55 | 74.51 | 81.75 |
| Large hemisphere (75$m$ radius) | 86.48 | 35.59 | 50.43 |
| NextBestView / Greedy | 90.99 | 78.18 | 84.10 |
| Roberts et al. [2017] | 88.19 | 77.17 | 82.31 |
| **Ours** | 94.12 | 82.28 | **87.80** |

Table 4. Quantitative comparison of the final reconstruction on the synthetic scene *Street* with a travel budget of 2700 m. Here we included the images from the initial scan when performing reconstruction. The best F-Score value is highlighted in bold.
Note that we had to limit the number of viewpoints used in our implementation of Roberts et al. [2017] to 4000. Otherwise Gurobi was not able to find any solution within a runtime of 60 minutes.

*5.5.1 Church.* Fig. 7 shows results for the *church* scene, acquired with a total of 160 images. The initial flight pattern uses 20 viewpoints arranged in an ellipse. Based on the initial reconstruction a viewpoint path with 140 viewpoints and a maximum flight time of 10 minutes was planned (see Fig. 1) and flown.

We also compare our final result with two baselines, a small ellipse (Baseline 1) and a large ellipse (Baseline 2) in In Fig. 7. Both patterns contain the same total of 160 images as ours. It is evident that the baselines are not able to recover the same amount of detail as ours. Furthermore, geometric concavities such as the portal are much better resolved by our method. We show a further view of the church in Fig. 8 where fine geometric details such as pane separators of the windows are visible. Note that we chose a constant number of viewpoints to provide a fair comparison. While the baseline methods could also record many more images this does often degrade performance as we demonstrate in the Appendix 9.3.

*5.5.2 Further results.* We show results from two further real-world experiments in Fig. 8, an *office* building with a very regular geometry and a *historic* building in a cluttered urban area showing the versatility of our method. For better comparison of details some areas are highlighted and enlarged. For the sake of brevity, we only show the best baseline for comparison.

Note the additional geometric details in our result compared to the baseline, in particular around windows, the stairs and sharp corners of the *historic* building. Even for the more regular and locally smooth geometry of the office building we see improved detail for the objects on the roof of the building and also surrounding the windows and along the roof line.

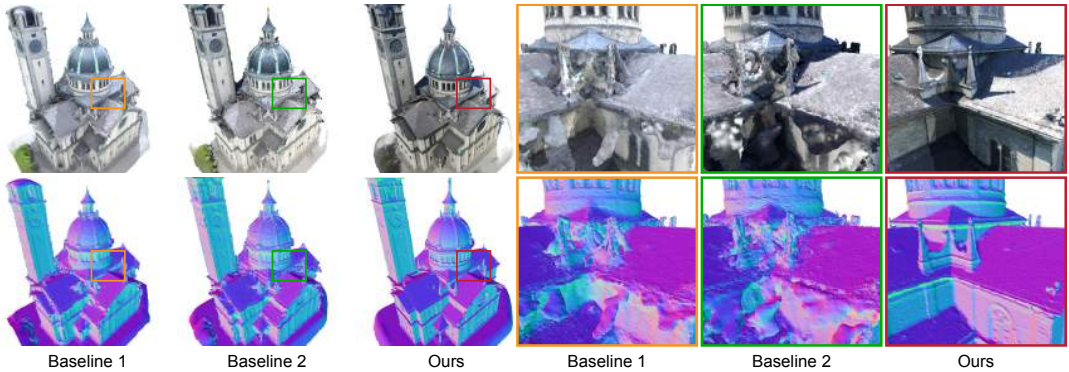

Baseline 1      Baseline 2      Ours      Baseline 1      Baseline 2      Ours

Fig. 7. Our result for the *church* scene. The top row shows color renderings and the bottom row shows normal renderings. In particular in the normal renderings a lot of detailed structures are visible that are smoothed out or corrupted in baseline reconstructions.

In both cases the total number of viewpoints was 80 and the maximum flight time was 10 minutes. We show the viewpoint paths in Fig. 9.

## 6 DISCUSSION AND FUTURE WORK

We have proposed a method for the automated planning of viewpoints and safe free-space trajectories for acquisition of aerial images for dense monocular reconstruction. At the heart of our method lies a mathematical optimization formulation that explicitly reasons about observed and free, occupied and unobserved space. A volumetric occupancy grid is used to compute information gained from observing parts of the scene which in turn is used to compute an optimized set of viewpoints. Furthermore, the framework incorporates domain knowledge about SfM & MVS pipelines into the formulation such as preferring fronto-parallel views and ensuring viewpoints can be matched well. An approximate camera-model that allows us to decouple viewpoints from each other allows for a sub-modular and hence efficient implementation.

We have demonstrated the versatility of our method using a number of varied scenes ranging from a renaissance church to a modern office building. Our method produces high-quality reconstructions and can recover high-fidelity geometric and texture detail, and is capable of resolving even difficult parts of buildings such as concave portals, roofs and other overhangs. Furthermore, using synthetic scenes we have shown quantitatively that our method outperforms baselines such as a hemisphere pattern and the prior state-of-the-art.

As mentioned before our resulting reconstructions are not perfect. This is due to several issues that are at play: First, a SfM/MVS reconstruction is a complicated process that is very hard to model and thus our planned sequence of viewpoints does not necessarily lead to the expected reconstruction outcome based on our modeling assumptions. Second, the sequence of viewpoints is limited by the travel budget which was set to 900 seconds in this case. Third, the dynamic range in the images is limited and surfaces under the porch are very dark and might even saturate for different viewpoints. Fourth, The set of possible viewpoints is finite due to sampling and we also require some distances from obstacles to prevent collisions (in this case 1$m$ distance from every mesh triangle in the scene, i.e. also 1$m$ above the ground). Finally, the dense point cloud from the MVS reconstruction might show inhomogeneous and low densities that lead to smooth arcs instead of sharp edges in the Poisson reconstruction of the mesh.

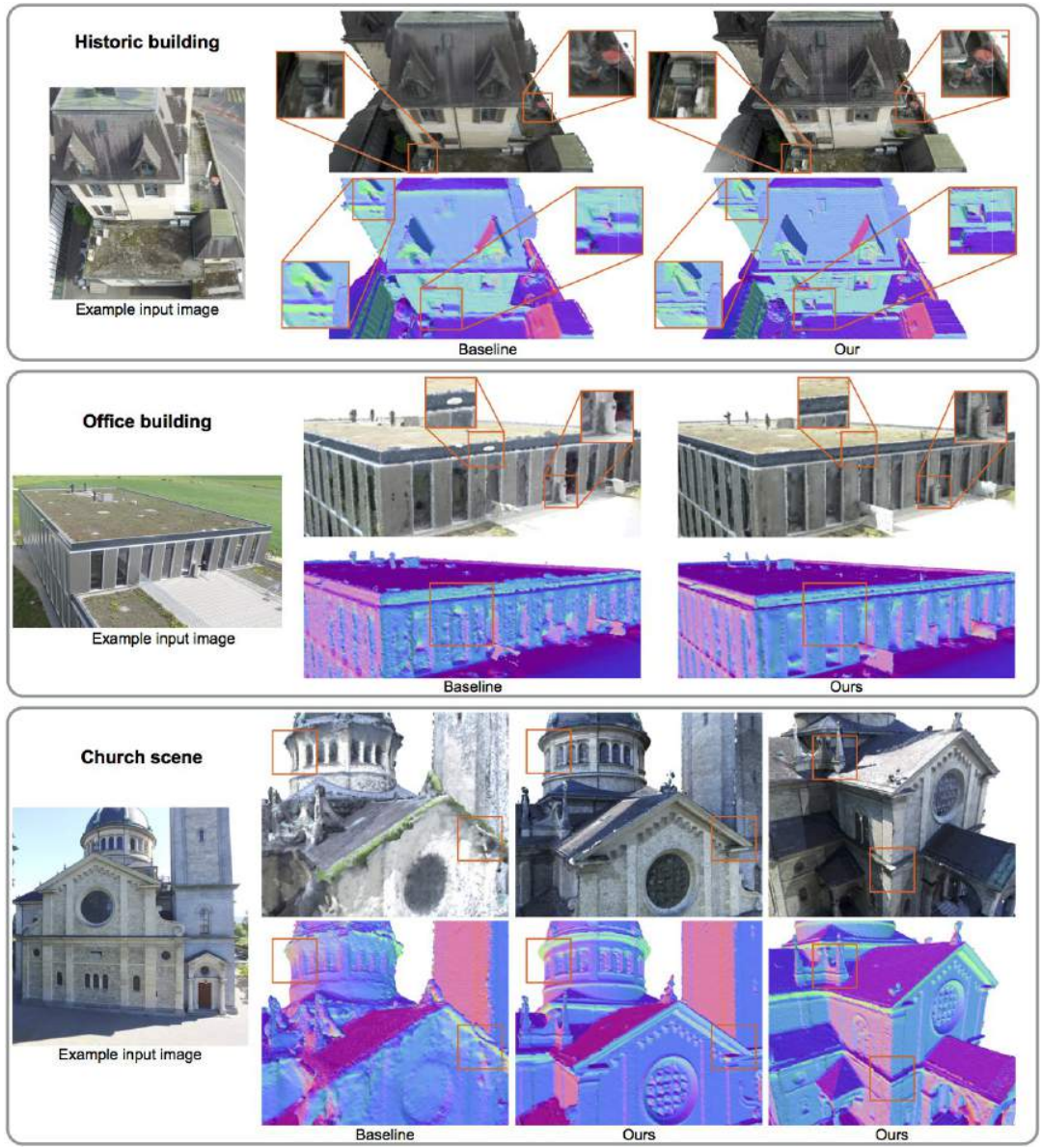

Fig. 8. Results of our method. Top row: *Historic building* scene. Middle row: *Office building*. Bottom row: *Church* scene. Note that in all cases, the baseline and our approach use the same number of view points; 80, 80, and 160 views, respectively.

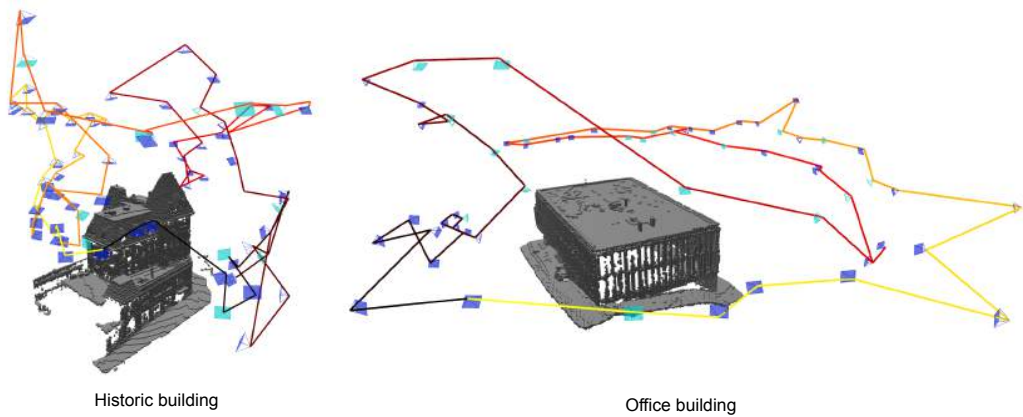

Fig. 9. Our planned viewpoint trajectories for the *Historic building* and the *Office building*. The viewpoints selected by our algorithm are rendered in blue and the additional sparse matching viewpoints necessary to ensure a successful reconstruction are shown in cyan. The edges are color-coded to signal MAV progress along the path from black to red.

A major limitation of our work is the need to split the scanning procedure into two phases, i.e. we first coarsly scan the scene from a safe altitude, then perform necessary computations and planning and finally follow the planned trajectory to capture images for the final high quality reconstruction. The time required for computation and planning (in our case this is around $30 - 60min$ dependent on the number of initial images and the desired quality level) often requires a second visit to the site which can be inconvenient or even prohibitive (if the scene or environmental conditions change quickly). While simple regular baseline patterns do not require a computation step the quality of the resulting reconstructions strongly depends on the structure of the scanned scene and for most buildings the need to fly at high altitudes (to avoid obstacles) results in steep viewing angles. We show a comparison of the required times in Sec. 9.4. Hence single-shot approaches with online planning or next-best-view selection would be desirable but may rely on robust and fast Structure from Motion and Multi View Stereo approaches both of which remain fruitful areas of future work.

For future work, we plan to further explore the iterative nature of our approach and are interested in incorporation of exploration strategies in order to scan even larger areas and to connect neighboring buildings. While we treat occupied space conservatively (i.e., even small noise in the depth maps is considered an obstacle), we can not guarantee that the initial scan succeeds to reconstruct thin structures such as powerlines in which cases collisions may be a possibility. Currently we offload this issue to the user and always fly with a safety pilot but think that this is a fruitful (and active) direction for robotics research. Finally, we think it would be an interesting direction of future work to combine our quality-focused reconstruction metric with an online depth map acquisition technique such as structured light scanners for online processing and planning.

## 7 IMPLEMENTATION DETAILS

### Occupancy Map

As is typical in occupancy mapping (Hornung et al. 2013; Thrun et al. 2005) we integrate depth maps into the occupancy map by updating the occupancy of each traversed voxel according to a beam-based inverse sensor model. We use an occupation probability for reflected beams of 0.7 and an occupation probability for transmitted beams of 0.4. Occupancy values are clamped to the range $[0.12, 0.97]$. A voxel is considered to be free if it's occupation probability is below $oc_{free} = 0.25$. We refer the reader to (Hornung et al. 2013) for more details.

### Recursive greedy method

In Alg. 1 the first step of the recursive procedure is to compute the middle viewpoint $V_m$ for the next recursion level. This viewpoint must be reachable with the current budget, i.e. the sum of the travel distance from $V_s$ to $V_m$ and from $V_m$ to $V_e$ must not exceed the budget. Additionally, from these viewpoints we want to select the one with the maximum information gain given the current viewpoints on the path $V$.

To speed up this computation we keep a separate list that stores for each viewpoint a tuple of the viewpoint index and the corresponding information gain given the current viewpoint set. Initially the list contains the information gain given an empty viewpoint set and is sorted with descending information gains. Each recursion step receives a copy of this list so that subsequent updates only effect the following recursion steps. When searching for the next middle viewpoint we iterate through the list starting from the beginning. If a viewpoint is not reachable from the current $V_s$ and $V_e$ we skip it. Otherwise we update the information gain value and perform a single bubble sort iteration (i.e. push the entry back in the list) until the list is sorted again. We continue this until we reach a viewpoint that does not need to be reordered. This is the reachable viewpoint with maximum information gain.

This procedure is well established for greedy submodular optimization (Krause and Golovin 2012). As the information gain is submodular it can only decrease (or stay equal) with additional viewpoints. Thus, after recomputing the information gain of a viewpoint it can only move backwards in the list. Importantly, if we recompute the information gain and the viewpoint keeps a higher information gain than the next viewpoint it must be the viewpoint with the maximum information gain as all information gains of subsequent viewpoints can only decrease (or stay equal). This lazy evaluation strategy typically results in only a couple of information gain computations that are necessary. We observed speed ups in the order of $10 - 20$ compared to a non-lazy evaluation.

### Dense Reconstruction Pipeline

We use standard structure from motion and multi-view stereo pipelines ( MVE (Fuhrmann et al. 2015) for synthetic experiments and Colmap (Schönberger and Frahm 2016; Schönberger et al. 2016) for outdoor experiments) to generate our models. We capture images by flying a DJI Matrice 100 drone with an onboard-computer. After landing, we perform structure from motion and multi-view stereo computations and filter the resulting depth-maps spatially by fusing the point-clouds, yielding a geo-referenced 3D model via incorporation of the image GPS coordinates. The depth maps are ray-casted into an occupancy map as described in Sec. 3. In cases where our model was initialized from a previous iterations, we fuse data into a single occupancy map. Finally, we generate a mesh from the dense point cloud using Screened Poisson Reconstruction (Kazhdan and Hoppe 2013).

**Trajectory flight**

Our planned trajectories are piecewise linear segments and the trajectories are flown via pure pursuit path tracking (Coulter 1992) with a desired velocity of $3m/s$ and a short lookahead distance of $2m$ leading to smooth motion. Note that we slow down at each viewpoint to take high resolution images without motion blur (i.e. when the next viewpoint is within a $3m$ distance we reduce the desired velocity to $0.5m/s$). We consider this in our viewpoint path optimization by adding a heuristic budget cost of $3s$ for each viewpoint.

## 8 ALGORITHMS FOR VIEWPOINT GRAPH GENERATION

The formal description for our viewpoint generation and motion computation can be found in Alg. 2 and Alg. 3, respectively. The procedure *addViewpointAtPosition* samples an orientation and adds the resulting viewpoint to the viewpoint graph and the exploration front. An orientation is sampled by sampling a yaw and a pitch angle (roll is fixed to zero). Both angles are sampled in the same manner by first computing the angle pointing towards the center of the region of interest and then adding an offset. The offset is sampled from a zero-mean gaussian with a standard deviation equal to the angular range resulting from the axis-aligned bounding box of the region of interest.

---

**Procedure** *generateViewpoints* **is**

 **Input:** List of initial viewpoints $V_0$
 **Input:** Minmum number of viewpoints min_viewpoints
 **Output:** List of viewpoints V
 exploration_front ← $V_0$;
 **while** |exploration_front| > 0 *or* |V| < min_viewpoints **do**
  **if** |exploration_front| > 0 **then**
   ref_vp ← random sample from V;
   exploration_front ← V \ ref_vp;
   **for** direction ← *6 axial directions* **do**
    step_size ← computeStepSize(ref_vp);
    p ← ref_vp.$p$ + step_size ∗ direction;
    addViewpointAtPosition(p);
   **end**
  **else**
   p ← random sample from allowed_space;
   addViewpointAtPosition(p);
  **end**
 **end**
**end**

**Algorithm 2:** Shown is the algorithm for generation the viewpoint candidate graph. The term *axial directions* refers to the $-x, +x, -y, +y, -z, +z$ directions.

---

## 9 ADDITIONAL RESULTS

### 9.1 Submodular optimization results

To estimate the effect of runtime length of Gurobi for the baseline method (Roberts et al. 2017) we ran Gurobi for selected travel budgets ($900m$, $1800m$) on the Grassland scene with small region of interest for 60 minutes. The resulting improvement in score was less than 1% in both cases.

**Procedure** *findMotions* **is**
   **Input:** List of viewpoints V
   **Input:** Number of neighbours to consider $K$
   **Output:** Dictionary $M : \{v_1, v_2\} \rightarrow m$ mapping two viewpoints to a motion $m$
   **for** $v_1 \in$ V **do**
      $N \leftarrow$ find $K$ nearest neighbours of $v_1$;
      **for** $v_2 \in N$ **do**
         $m \leftarrow$ findMotion($v_1, v_2$);
         **if** $m$ *is a motion* **then**
            $M \leftarrow M \cup \{\{v_1, v_2\} : m\}$;
         **end**
      **end**
   **end**
**end**

**Algorithm 3:** Shown is the algorithm for finding free-space motions between viewpoints in the viewpoint candidate graph.

### 9.2 Performance comparison when not including images from initial coarse scan

Here we report additional results when not including the images from the initial coarse scan for performing the 3D reconstruction. On the *Grasslands (small)* scene the F-score achieved by (Roberts et al. 2017) is $F = 71.74$ ($P = 91.12$, $R = 59.15$) whereas our method improves upon this with an F-score of $\mathbf{F} = \mathbf{76.95}$ ($P = 89.64$, $R = 67.41$). In Fig. 11 we qualitatively compare the reconstruction results, showing that our method can recover more surface details.

In Tab. 5, Tab. 6 and Tab. 7 we show quantitative results for our experiments when not including the images from the initial coarse scan.

The corresponding qualitative results are shown in Fig. 12.

| Method | Precision | Recall | F-Score |
|---|---|---|---|
| Small circle (35$m$ radius) | 77.29 | 13.26 | 22.64 |
| Large circle (70$m$ radius) | 72.55 | 3.92 | 7.44 |
| Small meander (70$m \times$ 70$m$) | 44.68 | 20.00 | 27.64 |
| Large meander (140$m \times$ 140$m$) | 43.60 | 20.57 | 27.95 |
| Small hemisphere (60$m$ radius) | 86.43 | 46.84 | 60.76 |
| Large hemisphere (75$m$ radius) | 79.35 | 48.77 | 60.41 |
| NextBestView | 91.53 | 22.08 | 35.58 |
| Roberts et al. (2017) | 88.61 | 46.44 | 60.94 |
| **Ours** | 89.72 | 50.70 | **64.79** |

Table 5. Quantitative comparison of the final reconstruction on the synthetic scene *Grass Lands* with a large region of interest and a travel budget of 1500$m$. Here we did not include the images from the initial scan when performing reconstruction. The best F-Score value is highlighted in bold.

### 9.3 Effect of number of viewpoints for simple baseline methods

In our experiments we use a fixed number of viewpoints for the static baseline methods which do not require an initial scan (i.e. circle and meander patterns). To show that our choice of a limited number of viewpoints does not lead to an unfair bias we compare reconstructions resulting from different numbers of viewpoints in Tab. 8. The results show that for all patterns but the large meander pattern (which performs poorly in both cases) the performance is lower when using the larger number of viewpoints.

| Method | Precision | Recall | F-Score |
|---|---|---|---|
| Small circle (50$m$ radius) | 86.98 | 15.88 | 26.86 |
| Large circle (75$m$ radius) | 83.21 | 6.50 | 12.06 |
| Small meander (85$m$ × 100$m$) | 61.48 | 11.34 | 19.15 |
| Large meander (100$m$ × 115$m$) | 66.09 | 10.60 | 18.26 |
| Small hemisphere (60$m$ radius) | 91.07 | 66.35 | 76.77 |
| Large hemisphere (75$m$ radius) | 90.19 | 44.03 | 59.17 |
| NextBestView | 94.58 | 63.67 | 76.32 |
| Roberts et al. (2017) | 92.93 | 72.15 | 81.23 |
| **Ours** | 93.59 | 74.30 | **82.84** |

Table 6. Quantitative comparison of the final reconstruction on the synthetic scene *Courtyard* with a travel budget of 2700 m. Here we did not include the images from the initial scan when performing reconstruction. The best F-Score value is highlighted in bold. Note that we had to reduce the number of viewpoints used in the method from (Roberts et al. 2017) to 3000. Otherwise Gurobi was not able to find any solution within a runtime of 60 minutes.

| Method | Precision | Recall | F-Score |
|---|---|---|---|
| Small circle (50$m$ radius) | 84.37 | 16.91 | 28.17 |
| Large circle (75$m$ radius) | 77.96 | 6.06 | 11.24 |
| Small meander (85$m$ × 100$m$) | 56.94 | 12.01 | 19.83 |
| Large meander (100$m$ × 115$m$) | 60.38 | 11.15 | 18.82 |
| Small hemisphere (60$m$ radius) | 90.38 | 76.52 | 82.87 |
| Large hemisphere (75$m$ radius) | 86.65 | 37.25 | 52.10 |
| NextBestView | 94.97 | 75.42 | 84.07 |
| Roberts et al. (2017) | 93.69 | 79.27 | 85.88 |
| **Ours** | 94.25 | 84.95 | **89.36** |

Table 7. Quantitative comparison of the final reconstruction on the synthetic scene *Street* with a travel budget of 2700 m. Here we did not include the images from the initial scan when performing reconstruction. The best F-Score value is highlighted in bold. Note that we had to reduce the number of viewpoints used in the method from Roberts et al. (2017) to 4000. Otherwise Gurobi was not able to find any solution within a runtime of 60 minutes.

| Method | Precision | Recall | F-Score |
|---|---|---|---|
| Small circle (25$m$ radius, 120 viewpoints) | 76.90 | 60.14 | 67.50 |
| Small circle (25$m$ radius, 240 viewpoints) | 52.72 | 48.69 | 50.63 |
| Large circle (50$m$ radius, 120 viewpoints) | 78.47 | 76.69 | 77.57 |
| Large circle (50$m$ radius, 240 viewpoints) | 50.27 | 61.48 | 55.31 |
| Small meander (50$m$ × 50$m$, 144 viewpoints) | 64.13 | 50.18 | 56.31 |
| Small meander (50$m$ × 50$m$, 256 viewpoints) | 51.11 | 39.10 | 44.31 |
| Large meander (75$m$ × 75$m$, 144 viewpoints) | 49.53 | 12.27 | 19.67 |
| Large meander (75$m$ × 75$m$, 256 viewpoints) | 47.20 | 14.01 | 21.60 |

Table 8. Quantitative comparison of the final reconstruction on the synthetic scene *Grass Lands* with a small region of interest. Shown are results for the one-shot baseline methods with different numbers of viewpoints. One can observe that the performance of the MVS reconstruction often decreases when more than 200 viewpoints are included.

### 9.4 Comparison of times for different methods

In table Table 9 we show approximate required times for the simple one-shot methods and the other methods requiring an initial flight. Note that the one-shot methods take less field-experiment time than the other methods. However, the resulting reconstructions also have a lower quality and just flying longer does not allow us to increase the quality as shown in the previous section.

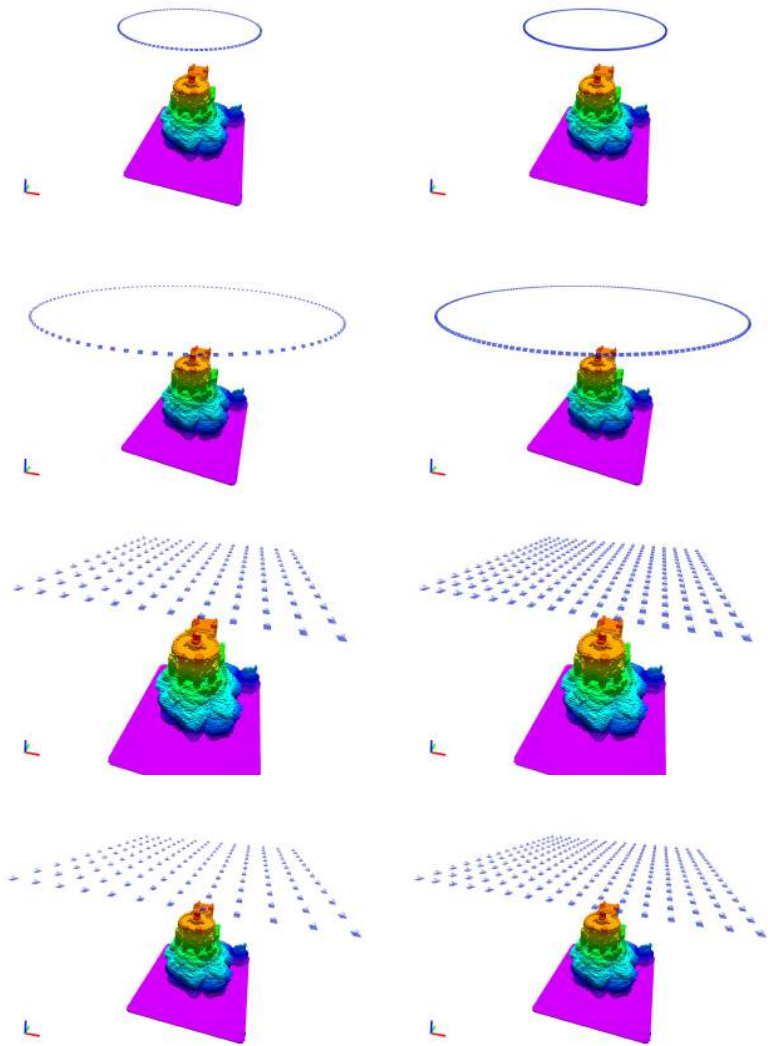

Fig. 10. Viewpoints used for comparison in Table 8. Shown from top to bottom: Circle with 25m radius, circle with 50m radius, Meander with 50m side length, meander with 75m side length. From left to right: Medium number of viewpoints, high number of viewpoints.

| Method | Flight time | Computation time | Total field experiment time |
|---|---|---|---|
| One-shot | $10min$ | $0min$ | $10min$ |
| With initial scan | $20min$ | $30 - 60min$ | $50 - 80min$ |

Table 9. *Flight and experiment times for the different methods compared in this paper.* Shown are estimated flight, computation and total field experiment times for the one-shot methods and the other methods requiring an initial scan. Note that the computation time depends on the specific scene. As we can see the one-shot method takes much less time.

## Grasslands (small)

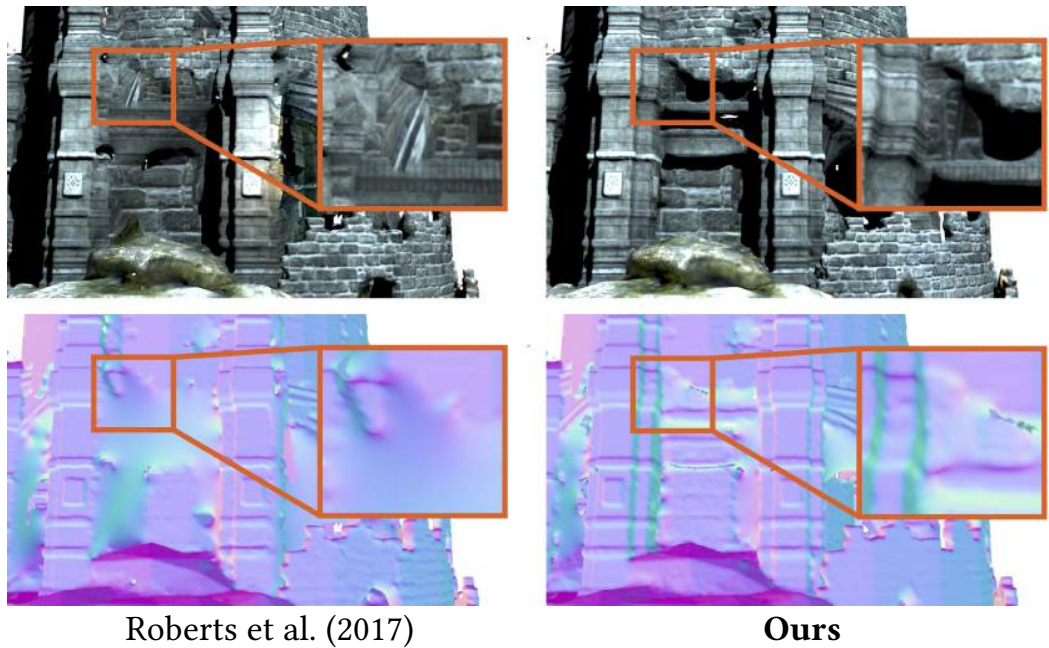

Roberts et al. (2017)                    **Ours**

Fig. 11. Qualitative comparison of (Roberts et al. 2017) and our method on an end-to-end dense reconstruction of the Grasslands (smal.) scene. Here we did not include the images from the initial scan when performing reconstruction. The first row shows the geometry without color whereas the second row shows the texture mapped reconstruction. The method of (Roberts et al. 2017) generates a viewpoint path that fails to capture some geometry information. Note that the missing geometry information also leads to distorted or erroneous textures. (Best viewed in color.)

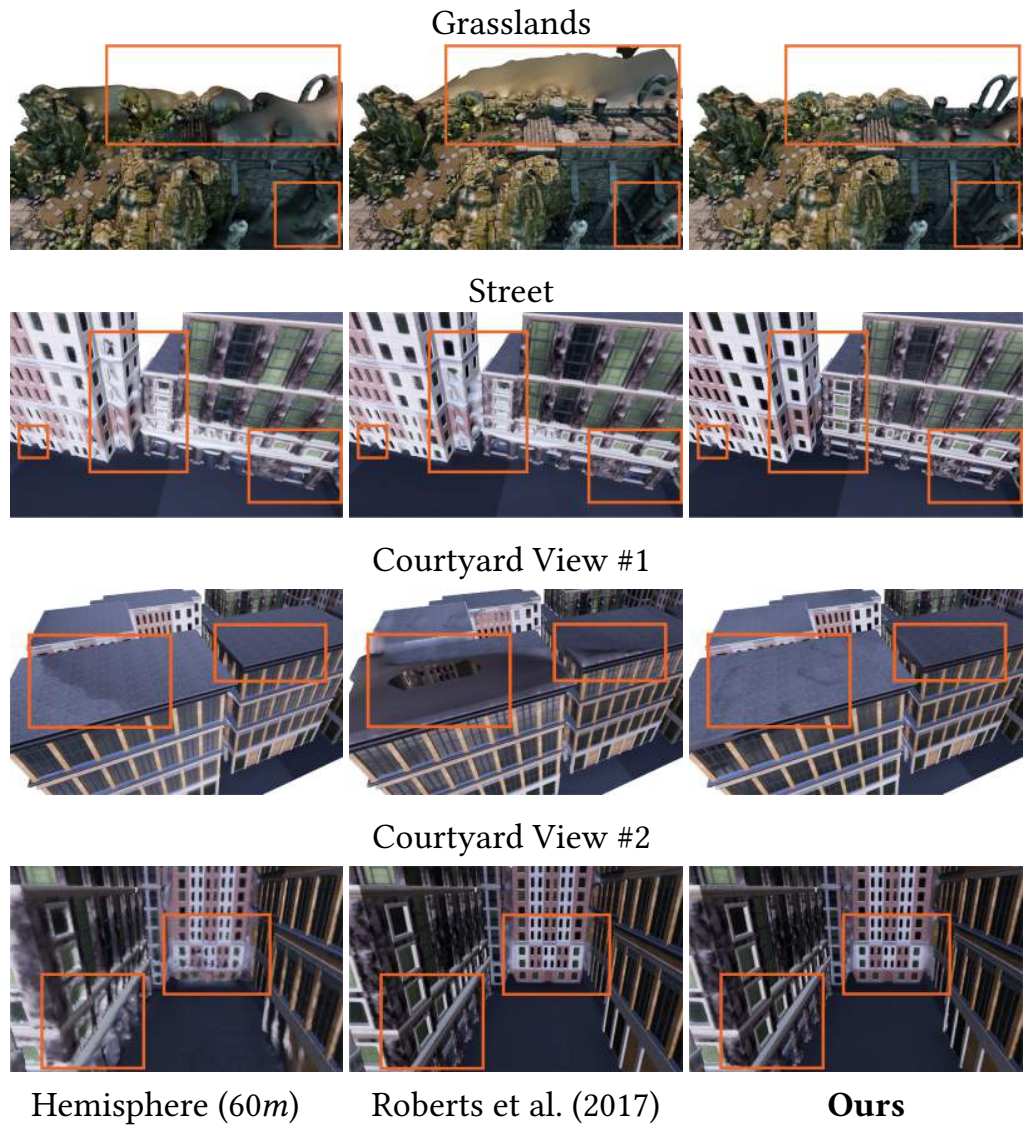

Fig. 12. Qualitative comparison of reconstructed meshes. For the Grasslands (large) scene we show results from a hemisphere pattern, (Roberts et al. 2017) and our method. Here we did not include the images from the initial scan when performing reconstruction. Both the hemisphere pattern and (Roberts et al. 2017) have missing points or lower point densities in some regions leading to bulging in the computed Poisson mesh. In contrast the reconstruction for our method shows much less bulging. For the *Courtyard* and *Street* scene we show results from a hemisphere pattern, the approach in (Roberts et al. 2017) and our method. Note that our method has more detailed texture and geometry in the lower facades of the building which requires viewpoints that move close enough but not exceeding the overall travel budget. (Best viewed in color and electronically.)